\documentclass[
  english,        
  font=times,     
  onecolumn,      
]{article}

\usepackage{authblk}
\usepackage{lineno}
\usepackage{placeins}
\usepackage{amsmath}
\usepackage{csquotes}
\usepackage{float}
\usepackage{booktabs}
\usepackage{multirow}
\usepackage{multicol}
\usepackage{siunitx}
\usepackage{soul}
\usepackage[utf8]{inputenc} 
\usepackage{colortbl} 
\usepackage{chemformula} 
\usepackage{xr} 
\makeatletter
\newcommand*{\addFileDependency}[1]{%
  \typeout{(#1)}%
  \@addtofilelist{#1}%
  \IfFileExists{#1}{}{\typeout{No file #1.}}%
}
\makeatother

\newcommand*{\myexternaldocument}[2][]{%
  \externaldocument[#1]{#2}%
  \addFileDependency{#2.tex}%
  \addFileDependency{#2.aux}%
}
\myexternaldocument{SI}

\DeclareUnicodeCharacter{2009}{\,} 
\DeclareSIUnit\angstrom{\text {Å}}
\DeclareSIUnit\rydberg{\text{Ry}}
\DeclareSIUnit\hartree{\text{Ha}}
\DeclareSIUnit\bohr{\text {\ensuremath {a}}_{0}} 

\usepackage{multirow}
\usepackage{geometry}
\usepackage{array} 
\usepackage{caption} 
\usepackage{makecell} 
\usepackage{hyperref}
\usepackage{float}      
\usepackage{chemformula} 
\usepackage{tabularx}   
\usepackage{placeins}   
\usepackage{graphicx}   
\usepackage{caption}    
\usepackage{amsmath}
\usepackage{amssymb}
\usepackage[capitalise, nameinlink]{cleveref} 
\usepackage{makecell}
\usepackage{comment}
\usepackage{bm}
\usepackage{filecontents}
\usepackage[natbib=true,
    backend=bibtex,
    style=numeric,
    sorting=none
]{biblatex}

\usepackage{etoolbox}

\title{A green solvent screening tool for emerging materials via uncertainty aware, transformer enhanced  transfer learning}

 \author[1,2]{Ioannis Kouroudis* $^{\dagger}$}
 
 \author[3,5]{Simon Ternes* $^{\dagger}$}
 
 \author[1,2]{Zhaosu Gu}
 \author[1,2]{Gohar Ali Siddiqui}
 \author[5]{Marina Ustinova}
  \author[4]{Angelo Lembo}
   \author[1,2]{Alessio Gagliardi*}
 \author[3,5]{Aldo Di Carlo}

\affil[1]{\small Chair of Simulation of Nanosystems for Energy Conversion, Department of Electrical Engineering, TUM School of Computation, Information and Technology, Atomistic Modeling Center (AMC)}
\affil[2]{\small Munich Data Science Institute (MDSI), Technical University of Munich, Hans-Piloty-Straße 1, 85748 Garching, Germany}

\affil[3]{Institute of Structure of Matter – National Research Council Rome (ISM-CNR), via del Fosso del Cavaliere 100, Rome, 00133, RM,Italy}

\affil[4]{Department of Chemical Science and Technologies, University of Rome Tor Vergata, Via della Ricerca Scientifica 1, 00133 Rome, Italy}

\affil[5]{Department of Electrical Engineering, University of Rome “Tor Vergata”, via del Politecnico 1, Rome, 00133,RM, Italy}

\date{ }

\begin{document}

\maketitle

\begin{flushleft}
\textsuperscript{$\dagger$}These authors contributed equally to this work.\\
*Corresponding Authors\\
Ioannis Kouroudis: \href{mailto:ioannis.kouroudis@tum.de}{ioannis.kouroudis@tum.de},\\
Simon Ternes: \href{mailto:ternes@ing.uniroma2.it}{ternes@ing.uniroma2.it},\\
Alessio Gagliardi: \href{mailto:alessio.gagliardi@tum.de} {alessio.gagliardi@tum.de}
\end{flushleft}

\begin{abstract}
Accurate prediction of solubility remains a central challenge across materials science and sustainable chemistry. In particular due to emerging technologies like organic and hybrid photovoltaics, batteries, and catalysis, solvent usage is expected to increase significantly within the coming years. Therefore, substituting solvents with greener alternatives is vital.  This is where machine learning can make a substantial impact. However, the limited  data on critical parameters of solubility significantly constraints machine learning efficacy.  In this work, we transfer a pre-trained foundational model on QM9 targets to our application with minimal data requirements. Additionally, the pipeline integrates uncertainty quantification, allowing the user to gauge the confidence of the predictions. As baseline, we succeed in predicting the Hansen solubility parameters and Dielectric Constant for which extensive databases exist. Importantly, we achieve high model performance on additional targets, such as Gutmann Donor and Acceptor numbers, where the available data is extremely limited. Overall, we augment data on solubility descriptors by up to two orders of magnitude with high quality predictions. For effective dissemination, we deploy easy-to-use, easily integrateable with high throughput labs, customizable tool for ranking and screening possible solvent substitutes.  Finally, we both rediscovered known green solvent alternatives and proposed new candidates proving its relevance for  finding eco-friendly solvents.

\end{abstract}

\section{Introduction}

Solvents or so-called volatile organic compounds (VOCs) are ubiquitous in the chemical, pharmaceutical, painting, cleaning and varnishing industries since the beginning of the industrial revolution\cite{calvo-floresGreenBioBasedSolvents2018}. Nowadays, on a global scale, over 20 million tons of solvents are used every year\cite{wintertonGreenSolventCritical2021}. Most recently, new energy-related emerging technologies such as batteries\cite{clementRecentAdvancesPrinted2022}, fuel cells\cite{choolaeiRecentAdvancesChallenges2023} and organic/hybrid photovoltaics (PVs)\cite{howardCoatedPrintedPerovskites2019} are starting to ramp-up solvent-intensive production. With significant market growth expected in these fields within the coming decade\cite{ThinFilmMateriala}, the environmental impact and safety concerns of the employed solvents become an urgent matter\cite{davidVolatileOrganicCompounds2021,pandeyReviewVolatileOrganic2018,pongboonkhumlarpHealthRiskAnalysis2022}. For example, the emerging field of perovskite PVs involves complex ionic solutions with organic and inorganic compounds and thus relies heavily on VOCs for the majority of state-of-the-art perovskite fabrication routines\cite{podapangiGreenSolventsMaterials2023,leeGreenSolventStrategies2025}. 

It is a complex evaluation process to weigh the advantages of employing a solvent for rolling-out a technology with its disadvantages\cite{wintertonGreenSolventCritical2021}. Especially for renewable energy-related technologies, potential benefits for human, animal and plant life must be contextualized with the associated impact of solvent waste, pollution and toxicity\cite{leeGreenSolventStrategies2025,zimmermannOrganicPhotovoltaicsPotential2012,vidalAssessingHealthEnvironmental2021}. There are, in turn, a multitude of strategies to reduce the negative impacts of solvent use in industry. These involve solvent recycling, the elimination of solvent-based processes or substitution of a solvent with an alternative of less concern\cite{wintertonGreenSolventCritical2021,podapangiGreenSolventsMaterials2023}. In the latter strategy, the term “green solvent” was introduced for chemicals which have reduced impact on the environment and less concern for operational employment. Detailed classification systems on the detrimental impacts of solvents on the environment and operators were developed, including criteria such i) possibility of incineration, ii) recycling, iii) bio-treatments and iv) the impact of VOC emissions in the atmosphere iiv) the aquatic impact iiiv) sourcing from sustainable materials, etc... These aspects are typically rated by an individual scoring system making “solvent greenness” comparable\cite{diorazioMoreHolisticFramework2016,capelloWhatGreenSolvent2007}. Thus, the selection of suitable green VOCs for industry is an interdisciplinary topic combining aspects of environmental science, material science and chemical engineering\cite{clarkeGreenSustainableSolvents2018,pratCHEM21SelectionGuide2016}. In every specific case, the solvent’s full environmental impact, its structural interaction with the targeted material system as well as its functionality in industrial processes such as coating, printing and drying must be studied\cite{capelloWhatGreenSolvent2007,byrneToolsTechniquesSolvent2016,clarkAlternativeSolventsShades2007}. 

One of the most important practical properties of a solvent is its ability to dissolve other chemicals (or, for liquid–liquid systems, its miscibility), which is governed by solvent–solute intermolecular interactions and, in some cases, specific complexation phenomena\cite{wintertonGreenSolventCritical2021}. While predicting solubility is a challenging task\cite{llompartWillWeEver2024,hopfingerFindingsChallengePredict2009}, a starkly simplified classification in "soluble" and "non-soluble" can be obtained by quantifying the strengths of certain chemical interactions in so-called solubility parameters\cite{bartonSolubilityParameters1975,vandykSolubilitySolvencySolubility1985,fernandesApplicationHansenSolubility2025}. The most widely-used framework, the three Hansen solubility parameters, quantify the strengths of dispersion forces, $\delta_d$, intermolecular forces, $\delta_p$, and hydrogen bonds, $\delta_h$ \cite{hansenHansenSolubilityParameters2007}. In this approach, the affinity between a solvent and a solute can be estimated by weighted distances between these three coordinates in the so-called “Hansen space”. Solubility is then classified by a distance threshold. However, Hansen parameters do not capture all interaction mechanisms that determine solubility\cite{fernandesApplicationHansenSolubility2025}. For example, they do not fully capture electrostatic, ionic, or acid-base interactions\cite{louwerseRevisitingHansenSolubility2017,lanComparingCorrelatingSolubility2014}.  In this context, Kamlet-Taft solvatochromic parameters  ($\alpha$, $\beta$, $\pi^*$)\cite{kamletLinearSolvationEnergy1983}, Gutmann donor/acceptor numbers (DN, AN)\cite{gutmannEmpiricalParametersDonor1976} as well as Dielectric constants ($\varepsilon_r$) can provide complementary information\cite{lanComparingCorrelatingSolubility2014,podapangiGreenSolventsMaterials2023}. However, the abundance of data on these parameters is not always given. While extensive databases with thousands of entries for measured and calculated $\varepsilon_r$ exist (just as for the Hansen parameters\cite{hansenHansenSolubilityParameters2007}), data on $\alpha$, $\beta$, $\pi^*$, DN and AN is very sparse---varying from several hundred to only tens of non-zero entries\cite{StenutzChemTables}. Thus, so far, the applicability of the latter parameters is limited to a very restricted subset of chemicals. An extensions to a large number of possible solvent substitutes is of high merit for discovery of alternative solvents.

Machine learning (ML) has an excellent track record in assisting material discovery. Its benefits are evident both in silico simulations \cite{kouroudis2023utilizing} as well as  real-world experiments \cite{mayr2022machine}. For instance, Henke \textit{et al.} and Lampe \textit{et al.} have used a  pipeline with Gaussian processes and Bayesian optimization on the synthesis process of hybrid perovskites \cite{henke2025synthesizer, lampe2023rapid}. Other examples are the exploration of new compositions of hybrid perovksite crystals \cite{pilania2016finding} and the identification of crystallization thresholds and luminescence properties  through image recognition  \cite{kirman2020machine,harth2023optoelectronic}. For the issue of prediction of solubility parameters addressed in this work, Hassan and Kazemi\cite{solubility_1}, Hu \textit{et al.}\cite{huPredictionDonorNumber2024} as well as Mahmood \textit{et al.} \cite{mahmood2023easy} have compared a range of simple machine learning models, using molecular descriptors as inputs, resulting in high-accuracy predictions. Similarly, Sanchez-Lengeling \textit{et al.} \cite{solubility_2} used simulation inputs to predict Hansen solubility parameters with Bayesian ML. Larsen \textit{et al.} combined similar predictions with a solvent-selection tool where simple environmental descriptors allow for identifying solvents with a lower environmental impact\cite{solubility_3}. These approaches were shown to yield consistent results within the investigated dataset. However, for material discovery, robust models that prioritize generalization over performance are needed. Therefore, to prevent over-fitting and foster generalization, we employ transfer learning on foundational models, coupled with uncertainty quantification and tested by n-fold cross-validation. Large foundational models, such as MACE \cite{MACE}, DimeNet \cite{dimenet} and NequIP \cite{batzner20223} are based on complex architectures involving millions of parameters trained on large datasets to predict a large variety of molecular properties at high fidelity. However, their accuracy sharply diminishes with the number of available training data. This poses a problem for their application to many experimental properties where only a relatively small number of data points are available. Transfer learning provides an elegant solution to this issue by training a slim add-on model on the outputs of foundational model to adapt them to a new domain. Some prominent examples are transfer learning of crystallization processes\cite{CRYSTAL_TRANSFER} and molecular properties \cite{yamada2019predicting}. However, despite these successes, transfer learning is not yet widely adopted in materials discovery, likely because its effective use requires careful alignment of tasks and objectives. To address this issue, Kolluru \textit{et al.} presented a versatile and facile transfer learning method by training transformers on the latent space of foundational models \cite{TAAG}.

In this work, we introduce a pipeline that is constructed on the above foundations, adding complementary information with molecular descriptors. We further extend the state-of-the-art by uncertainty quantification via Gaussian processes in the final layer of the pipeline. Gaussian processes have shown high promise in material applications when combined with graph neural networks\cite{kouroudis2025augur}. For the present work, they offer two main advantages: 1) They offer a confidence interval of the predictions to experimentalists that can be leveraged for incentivizing new experiments and 2) they provide robust performance even on sparse data effectively preventing overfitting by means of the prior distribution. As a main result of this work, we succeed in predicting vectors of solubility parameters\footnote{Some of these predictors are correlated, but they can still yield valuable complementary information.}, ($\delta_d$, $\delta_p$, $\delta_h$, $\varepsilon_r$, DN, AN, $\alpha$, $\beta$, $\pi^*$), on a database of over 5000 VOCs\cite{CompsolDatabaseRevised2024}. These vectors are normalized component-wise and compared by measuring the distance to a reference solvent to rank the similarity between the reference and green alternatives. To maximize the dissemination of our work, we deploy a ready-to-use solvent selection tool conveniently performing this ranking with custom weights. While regulations on solvent use in industry for reducing environmental impact and ensuring healthy work conditions become more rigorous, finding a "greener" solvent alternative could be the missing cornerstone for rolling out emerging revolutionary technologies to the commercial market at scale.

\section{Methods and Models}

\subsection*{Solvent similarity ranking for solubility prediction}\label{sec:ranking}

Solubility refers to the maximum concentration of a solute that can be homogeneously mixed in solution with a solvent due to the chemical interactions between solvent and solute\cite{iupac_solubility}. Predicting solubility quantitatively is a very challenging task because complex thermodynamics of solid-state stability and solute-solvent interactions over multiple length scales are involved and experimental uncertainty is high\cite{llompartWillWeEver2024,hopfingerFindingsChallengePredict2009}. A phenomenological, widely-used approach to profile the tendency of certain chemicals to undergo chemical interactions decisive for solubility are so-called 'solubility parameters' \cite{bartonSolubilityParameters1975,vandykSolubilitySolvencySolubility1985,fernandesApplicationHansenSolubility2025}. The most popular ones, the Hansen solubility parameters, divide the energy of vaporization of a molecule needed for breaking chemical bonds into three parts: 1) molecular atomic dispersion forces 2) molecular dipole-permanent forces and 3) hydrogen bonding (electron exchange)\cite{hansenHansenSolubilityParameters2007}. The associated energies are reflected in the parameters $\delta_d$, $\delta_p$ and $\delta_h$ measured in units of MPa$^{0.5}$. Similarity of chemical interaction profiles can then be measured as (weighted) distances of the Hansen parameters in the Hansen space. The question of solubility---starkly simplified down to a binary categorization of "soluble" (or "miscible") and "non-soluble" ("non-miscible")---is then addressed by defining a threshold of Hansen Distance between solute and solvent, which gives correct rough estimates in many cases\cite{venkatramCriticalAssessmentHildebrand2019}. However, data on Hansen solubility parameters for solutes of typical applications such as perovskite PVs and polymers is rare (and often not statistically validated)\cite{venkatramCriticalAssessmentHildebrand2019,babaeiHansenTheoryApplied2018}, while data of solubility parameters of solvents is more extensive\cite{hansenHansenSolubilityParameters2007}. Therefore, for the issue of solvent substitution, rather than calculating distances between solute and solvents, it is a more robust approach to compare the reference solvent to be substituted with another solvent. This introduces a maximum statistical error factor of two if the new solvent and the solute lie in opposite positions to each other in Hansen space (with the reference solvent situated in between). However, it eliminates the uncertainty of choosing the correct solubility parameters for the solute. Ranking for smallest distance and considering the possibility that statistical deviation can occur in a beneficial direction, comparing the solubility parameters of two solvents is a good option for solvent substitution. To increase the chances of success, one can use multiple  reference solvents that are known to dissolve the material system of interest. Experimentally, this approach could be validated by testing the miscibility of the reference solvent with the screened solvent candidates.

A fundamental limitation in prediction power of the Hansen solubility parameters is imposed by their definition. Importantly, they do not accurately capture donor-acceptor complexes as well as electrostatic interactions due to ionic charge\cite{louwerseRevisitingHansenSolubility2017,kobayashiDeterminationHansenSolubility2020,holzweberMutualLewisAcid2013}. These interactions can be crucial for solubility in solutions of ionic and acidic/basic solution, impacting for instance the applications of hybrid perovskite PVs\cite{podapangiGreenSolventsMaterials2023} and battery electrolytes\cite{zhouUnderstandingApplyingDonor2024}. To address the issue of predicting donor-acceptor complexes, Gutmann proposed to measure the "donor number" (DN) and the "acceptor number" (AN) as quantitative measures based on the negative enthalpy of formation of the 1:1 Lewis acid–base adduct between the standard Lewis acid $\mathrm{  SbCl}_5$ and Triethylphosphine oxide ($\mathrm{Et_3PO}$, TEPO)\cite{gutmannEmpiricalParametersDonor1976}. Historically, they were first determined by calorimetric measurements, but are nowadays mostly measured by NMR spectroscopy\cite{schmeisserGutmannDonorAcceptor2012}. Seven years later, Kamlet and Taft introduced parameters based on solvatochromic shifts of molecules in solutions measured spectroscopically\cite{kamletLinearSolvationEnergy1983}: The  hydrogen bond donor acidity parameter, $\alpha$, the hydrogen bond acceptor basicity, $\beta$ and the dipolarity/polarizabilty parameter $\pi^{*}$.  They are derived from spectral shifts in UV/VIS absorptions spectra or fluorescence spectra due to dissolution of the chemicals. Another important parameter for predicting electro-static effects is the dielectric constant of the solvents, $\varepsilon_r$. Note that all above solubility parameters can be correlated for some classes of chemicals\cite{lanComparingCorrelatingSolubility2014, parutaCorrelationSolubilityParameters1962, shiRelationshipHansenSolubility2007,waghorneSolventAcidityBasicity2024}, however it is unlikely any correlation would hold over the whole range of organic VOC molecules. Further note that different choices and weighting of these parameters should be applied depending on the target application. For example, hybrid perovskite solutions are ionic and well-known for the formation of Lewis-Base:Adducts with Dimethylsulfoxide (DMSO)\cite{liRationalDesignLewis2023,leeLewisAcidBase2016}. They further include small organic molecules such as methylammonium and formamidinium. Thus, for hybrid perovskite ionic solutions,  the parameter set ($\delta_d$,$\delta_p$,$\delta_h$, DN, $\beta$,  DC)  should be decisive for selecting a suitable solvent. For organic solar cells, on the contrary, the three Hansen solubility parameters might be sufficient for solvent substitution. To account for the varying importance of solubility parameters depending on the application, we propose a weighted Euclidean distance, $\hat{R}_{S_i,S_\mathrm{ref}}^2$, for solvent ranking as a generalization of Hansen distance to the additional parameters (zero weights can be given if parameters should not be considered).  For ranking potential candidate for solvent substitution, $S_i$, the distance between every candidate to the reference solvent $S_\mathrm{ref}$ to be substituted is calculated as

\begin{equation}
\hat{R}_{S_i,S_\mathrm{ref}}^2 = \sum _{n} w_n^2\left( \hat{p}_{n,\mathrm{ref}}-\hat{p}_{n,i} \right)^2,
\label{eq:dist}
\end{equation}

where we consider $n=9$ solubility parameters in total and the hat notation indicates a component-wise median absolute deviation (MAD) on the predicted range of the respective parameter $p_n$. The normalization is important to remove the impact of different units as well as predicted ranges of the parameters. As discussed above, the weights can be adjusted according to the chosen application. Another possible choice of weights $w_i$ for general solvent substitution is given by the $r^2$ values obtained by our ML pipeline to prioritize more accurately predicted parameters. This is what we use as a standard setting in the Results section. As another customization choice, we provide alternative metrics such as the Mahalanobis distance and Kullback–Leibler divergence with the goal to leverage the predicted uncertainties, as well.  All of these choices are integrated in the solvent screening tool discussed below.

\subsection*{Data mining}
For the application of ML on solubility parameters, the availability of data on either of these parameters plays an important role. While extensive databases with thousands of entries exist on Hansen parameters and $\varepsilon _r$ \cite{hansenHansenSolubilityParameters2007,}, data of the other parameters mentioned above is less abundant. To our knowledge, data is restricted to several hundred entries for AN, DN and $\beta$ down to only tens of non-zero entries for $\alpha$ and $\pi^{*}$\cite{StenutzChemTables,huPredictionDonorNumber2024}.  The main sources of data for this work was the website by R. Stenutz\cite{StenutzChemTables} accessed in June 2024,  and the book Hansen Solubility parameters a Users Handbook by Charles M. Hansen\cite{hansenHansenSolubilityParameters2007}. We have further consulted the work of Hu et al. for Gutmann donor and Acceptor numbers \cite{huPredictionDonorNumber2024}. Information on the molecular structure and information about the H-Symbols (GHS system) was pulled from PubChem\cite{pubchemPubChem}.  Note that these sources of digital data can be substituted easily in the future if inconsistencies become apparent or more extensive datasets are found. The tool will be updated accordingly. The 5000 VOC candidates were extracted from the CompSol database\cite{CompsolDatabaseRevised2024} (in fact not all of the candidates are liquid in room temperature, but dissolution at elevated temperatures might be an option for some technologies). For the Dielectric Constant parameter we filtered out values above 100 as solvent materials are not expected to have higher values and therefore, this points had a high probability of being erroneously measured.

\subsection*{Construction of Models and Algorithms}

In our study, we chose to use DimeNet (Directional Message Passing Neural Network) \cite{dimenet} as the foundational model for predicting solubility parameters of the investigated solvent molecules. We note that more foundational models such as MACE\cite{MACE} and NequIP\cite{batzner20223} have been developed since the start of the project. DimeNet can be substituted in future works without loss of generality or applicability of our pipeline. The SMILES notation, atom positions and atomic numbers for each molecule are pulled from PubChem via their IUPAC-Name. SMILES is used to filter the molecules for the occurrence of atomic groups that would set the expected solubility parameter to zero\cite{mathieuPencilPaperEstimation2018,gutmannEmpiricalParametersDonor1976,schmeisserGutmannDonorAcceptor2012,burkeSolubilityParametersTheory1984} for AN and $\alpha$. This alleviates some of the learning load of the subsequent model. In case of non-zero values, the atom positions and atomic numbers are used as input to the DimeNet model. The DimeNet model in its default architecture outputs one value per molecule. This constrains transfer learning, as using the output directly is analogous to using the relevant predicted property and not the reach matrix of molecular descriptors learned for this task.  The penultimate layer of DimeNet however contains a matrix M of dimensions $\mathbb{R}^{n\times m}$ where $n$ is the number of atoms in the molecule and $m$ the number of pre-chosen interaction blocks. We perform a column-wise summation of this matrix thus acquiring a fixed size  vector for our molecules. This vector contains a higher information density than the one-dimensional final output, thereby making it  more conducive to transferability. In \texttt{pytorch}, there are 11 readily available pre-trained DimeNet models, each on a different QM9 property. The QM9 database comprises approximately 134,000 small organic molecules combining 3D geometries with Density Functional Theory-calculated quantum-chemical properties such as dipole moments, rotational constants and isotropic polarizability, etc... \cite{qm9}. We optimize the choice of pre-trained model based on the prediction accuracy of the full pipeline as a hyperparamaeter (see \cref{tab:hyperparameters} for a full list of hyperparameters). In addition to the vector obtained from DimeNet, we use an embedding vector, $\mathbf{e}_n$, based on descriptors with dimensionality $n$. 
The descriptors were determined following the methodology of Mahmood \textit{et al.} \cite{mahmood2023easy}: At first, we append molecular descriptors to the graph dataset from the OCHEM  database\cite{ochem}. We then obtain the SMILES fingerprint using RDkit and search for it in the OCHEM database. Once matched, we remove descriptors which have too sparse values or low variance. In the end we select descriptors with low self-correlation and correlation with the target of higher than 0.5. 

\begin{figure}[h]
\centering

\includegraphics[width=1\textwidth]{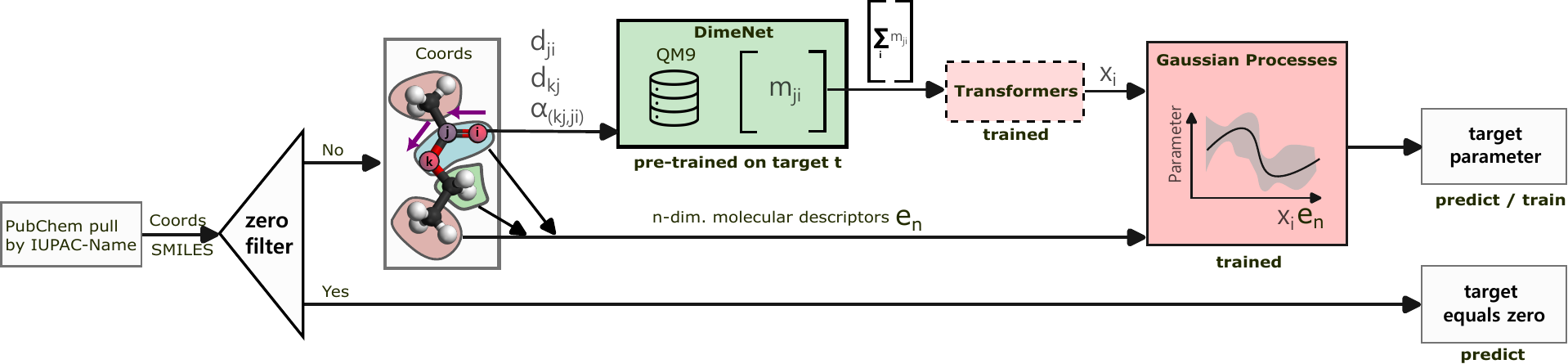}

\caption{Machine Learning pipeline used in this work. A SMILES-based filter firstly identifies if the chemical is expected to have a non-zero target parameter. For non-zero parameters, the foundational model DimeNet is used on the molecular coordinates pulled from PubChem and complemented with n-dimensional molecular descriptors $\mathbf{e}_n$. A Gaussian Processes regressor is then trained on a vector obtained by summation over the columns of the matrix M of the last DimeNet layer along with the molecular descriptors, $\mathbf{e}_n$, to predict the targeted solubility parameter.  For some parameters, it was beneficial to process the DimeNet output with a transformer layer for attention based value adjustment. }
\label{fig:main_schematic}
\end{figure}

This step allows the DimeNet outputs and the descriptors to adjust each other, allowing for a result that is larger then the sum of its parts. Finally, the transformer output is used as input for a Gaussian Process (GPs) regressor which provides a robust prediction as well as a vital uncertainty quantification.  The transformers and GPs are trained simultaneously towards the minimization of the negative log likelihood of the predictions. The DimeNet weights remain frozen, leveraging the full benefits of the pre-training. A clear indication of the effects of each component can be see in section \cref{sec:comparison} where an ablation study is performed and models with different components are compared to each other.  A succinct representation of our complete pipeline can be seen in \cref{fig:main_schematic}

The scarcity of the available experimental data of Kamlet-Taft $\beta$, $\alpha$, $\pi*$, Gutmann AN and Gutmann DN numbers discussed in the last section poses a challenge for the application of predictive ML. A possible mitigation strategy for low sample numbers is to employ models that propagate associated uncertainty with the predicted values. In this way, they are intrinsically regularized against over-fitting by quantifying the gaps in the knowledge of training data. This is the reason why  the last layer of our model incorporates the data-driven stochastic algorithm, Gaussian processes (GPs). The algorithm models predictions as the posterior of the Bayes formula, while the prior and the likelihood are modeled as Gaussian distributions whose parameters are optimized based on the already measured samples \cite{rasmussen}. GPs combine robustness to overfitting and the well-documented accuracy of global kernel methods (a more detailed description can be found in
\cite{williams2006}). Note that with state-of-the-art algorithms implemented in \texttt{gpytorch} \cite{gpfast1, gpfast2}, GPs can easily handle datasets of a million samples with very limited approximations. This why GPs work both on scarce as well as large datasets.

\begin{table}[H]
\centering
\begin{tabular}{ll}
\toprule
\textbf{Hyperparameter} & \textbf{Value} \\
\midrule
GP mean & Linear, Constant, Zero \\
Number of gaussian mixtures in kernel & 2, 4, 6 \\
pre-trained DimeNet target & 0, 1, 2, 3, 5, 6, 7, 8, 9, 10, 11 \\
Property pre-processing & Standardization, Normalization \\
Transformer heads & 1, size of input vector \\
\bottomrule
\end{tabular}
\caption{Hyperparameter values that were explored for every predicted parameter.  }\label{tab:hyperparameters}
\end{table}


\subsection*{Solvent Screening Tool}\label{sec:screening}
As introduced above, an exact definition of the colloquially used term “green” is challenging, although a matrix of factors can be considered that make “solvent greenness” comparable\cite{diorazioMoreHolisticFramework2016}. Some of these factors are mirrored in the GHS hazard system that is used on a daily basis by chemists and engineers to assess pre-cautionary measures before a solvent is employed\cite{moritaExpertReviewGHS2011}. The cataloged GHS characteristics were designed to guide short-term decision making in how to handle solvents. Therefore, they do not necessarily represent long-term characteristics such as the full environment impact and life cycle of the industrial use of a certain chemical, starting from the production of the solvent from natural resources, through its mixture with other materials and possible contamination to the recycling and/or release to the environment after waste treatment\cite{clarkeGreenSustainableSolvents2018,capelloWhatGreenSolvent2007,FrameworkSolventRecovery2019}. To give an extreme example, the 'greenest of all solvents', water, is still a limited resource in many areas of the word and contaminated water cannot be released into the environment without treatment\cite{richardsonWaterAnalysisEmerging2011}. When selecting a solvent for an industrial process, all these factors must be taken into account for determining its true environmental impact. Still, the functionality of a solvent for a certain application also plays into the balance of this cost/benefit analysis. As inscribed in the name itself, the functionality of a solvent is, first and foremost, determined by their ability to dissolve or mix with a certain class of materials\cite{wintertonGreenSolventCritical2021}. At the same time, its secondary solvent properties such as the viscosity, the boiling point, the diffusion coefficients, stability under stress factors, chemical reaction pathways can play an important role\cite{enekvistComputeraidedDesignSolvent2022,pilonDevelopmentSolventSustainability2024}. 
For the tool developed in this work, we focus on solubility as primary performance indicator. Further, we apply very rudimentary filters to identify possible “green solvents”. By choice of the user, these filters simply exclude any H2xx (physical hazards), any H3xx (health-related hazards) and any H4xx (environmental hazards) symbols to assess whether a chemical is a “green” alternative. Note that more detailed filters of H-Symbols could have been added to the tool. This was deliberately ruled out to avoid users relying on the provided information instead of a proper safety data sheet. However, when working on the raw data on their own risk, users can implement these filters themselves. The solvent selection tool was deployed as a lightweight \texttt{streamlit} application. The main idea is that any reference solvent to be replaced can be chosen and the database is then sorted according to the smallest distance to the reference solvent according to \cref{eq:dist}.  In the tool, the user can select any solvent in the database that they want to find a similarity for and get auto-generated tables like these ones in csv, Excel or \LaTeX format. The weights of either predictor can be either chosen by the $r^2$ values of the predictions or custom selected. A further customization is given by choosing between three different distance metrics. In the Euclidean distance we provide its propagated uncertainty in a separate column, while Mahalanobis distance and Kullback–Leibler divergence already incorporate the uncertainties predicted by our pipeline. The tool is available under \url{https://kouter.streamlit.app/} (Github page \url{https://github.com/cassimon/kouter})\footnote{Tool will be released upon acceptance of manuscript}. Note further that the Hazard symbols were mined from PubChem and thus errors in the screening are to be expected\footnote{The tool should never be used for any practical laboratory decisions and only for research purposes (as is written in its disclaimer)}. The benefit of the tool as compared to previous works is that it is customizable to the needs of different applications and offers predictions over a wide range of over 5000 VOCs. To customize the tool for their purposes, users can select custom weights to choose the exact prioritization of interactions that matters for their particular material system. Further, the user can exclude chemicals of high uncertainty for conservative solvent substitutions or include even highly uncertain ones for a broader exploration of the chemical space. 

\section{Results \& Discussion}
\subsection*{Performance of Machine Learning Pipeline} \label{sec:comparison}

\begin{figure}[!h]
\centering
\includegraphics[width=1\textwidth]{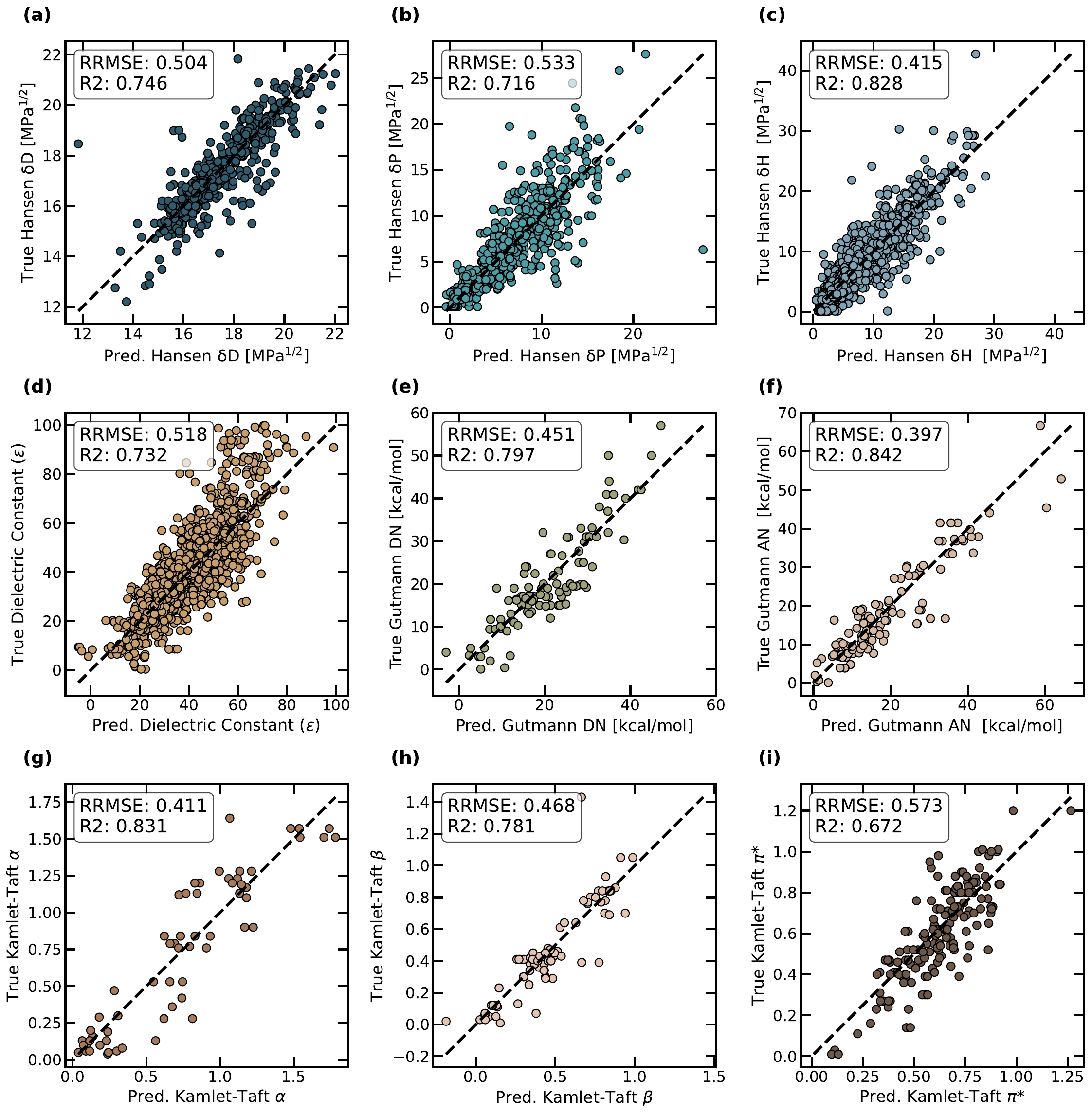}
\caption{Regression curves resulting from the pipeline shown in  \cref{fig:main_schematic}, for all the relevant parameters for the solubility. The results are generating by 5 fold validation for the relatively large datasets and 10 fold validations for the smaller ones. All the presented results are predictions on the test set, i.e. the data not seen by the model during training.   }
\label{fig:main_results}
\end{figure}

The combined predictions on the test sets are presented in \cref{fig:main_results} for all solubility parameters. The results presented are the combined test set predictions of an $n$-fold validation scheme ($n=5$ for large and medium datasets and $n=10$ for sparse datasets). Thus, we make certain that our model performs consistently without reporting a fortunate data split or weight initialization. In this way, we confidently expect our model to generalize well even on unseen data. Comparing the predictions on large datasets (several $1000$ points) and medium datasets (several $100$ points) to very small dataset (less than $50$ points), we find no significant decay in performance. This data efficiency underlines the strengths of transfer learning with foundational models that require fewer training points to reach decent performances as compared to models that need to be trained from scratch. In addition, the majority of predictions are symmetric to the line of perfect agreement (dashed line), indicating that the prediction errors are largely unbiased. The $\delta_d$ and $\delta_h$ parameters exhibit a slight systematic deviation in the high and low value regime, likely caused by the scarcity of training data in these regions. 

\begin{figure}[!h]
    \centering
    \includegraphics[width=\linewidth]{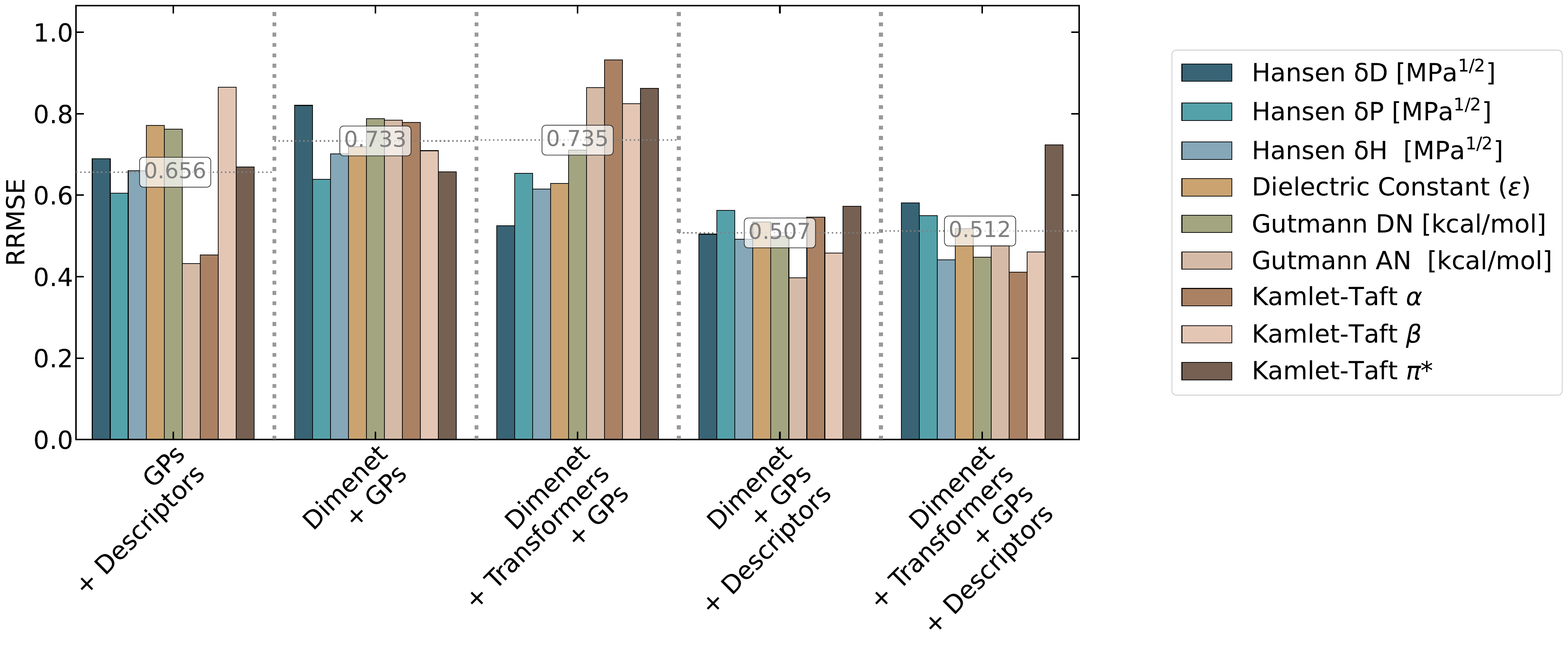}
    \caption{RRMSE for every predicted solubility parameter omitting different components of the pipeline shown before in \cref{fig:main_schematic} keeping  Gaussian Processes as a final layer. The baseline model relying only on molecular Descriptors as inputs, reaches a decent average performance. While DimeNet achieves a similar performance to the GPs and molecular descriptors,  a significant improvement is achieved when integrating DimeNet outputs with descriptors.  A transformer layer gives further improvement for some parameters, where attention is beneficial. }
    \label{fig:comparison_soa}
\end{figure}

In pursuit of streamlining our pipeline, we tested the performance impact of every element. In detail, we test the incomplete pipeline where we omit and combine elements in all relevant configurations. On every test, we compare the relative root mean square error (RRMSE) performances on the $n$-fold validation with the full pipeline. For a fair comparison, a hyperparameter sweep was performed every time (see \cref{tab:hyperparameters} for an overview of hyerparameters). The results are depicted in \cref{fig:comparison_soa}. The state-of-the-art baseline, GPs and descriptors, achieves a decent performance for our dataset with a mean RRMSE of $0.656$. In comparison, using the DimeNet outputs alone does not yield a performance difference. However, concatenating the descriptors with the DimeNet outputs and using the resulting vector as an input to GPs, we obtain a significant  improvement across all parameters. This is a crucial result as it points to the two distinct information sources. The descriptors encompass the distillation of human chemical intuition and are therefore of great importance for the prediction performance. Without descriptors, the rest of the pipeline has to compensate for the knowledge gap. 
Including them however allows the model to leverage data driven learning for learning the remaining mechanisms not captured by expert knowledge. 
In this way, we combine the advantages of expert knowledge and data driven discovery.  This is evident in \cref{fig:comparison_soa} where the models that contain both the foundational model and the descriptors significantly outperform all the others. The addition of a processing step of the concatenated vector through a transformer layer further improves the result on some of the parameters. This becomes more evident in the low data regime, as the addition of another layer of complexity increases the data load required. Therefore, transformers add consistent improvement in cases of large data sets and only provide better performance in small data sets if the DimeNet and descriptor variables are strongly correlated to the output in ways that the attention mechanism can decipher.

With this in consideration, our pipeline, with or without transformers significantly outperforms the GPs with descriptors by a margin of at least $31 \%$, with some properties improving by as much as $85\%$.
Further, in comparison to other state of the art approaches, our pipeline also provides a crucial uncertainty quantification, which allows for a more thorough understanding of the prediction space. 


Finally, to reference our results to the state-of-the-art, we summarize the performance of our full pipeline along with a baseline model of GPs and molecular descriptors\cite{mahmood2023easy} for every predicted solubility parameter in \cref{tab:compariosn_models}.  For the large and medium data set size, our final pipeline outperforms the baseline model by approximately 0.25 RRMSE (~50\% relative improvement).  
The performance improvement decreases on some of the small datasets. In all probability this is because the molecules included in these datasets are not overly chemically diverse, thus making the prediction of the test set relatively easier. Nevertheless, even on those datasets our pipeline provides a meaningful and measurable improvement.
To sum up, we find that our pipeline excels both on large and on small datasets, where conventional approaches struggle. Even on large datasets, it performs on par with conventional approaches. Our versatile and adaptable pipeline is therefore a worthwhile addition to the state-of-the-art. 

\begin{table}[h!]

\centering
\begin{tabular}{cccc}
\toprule
\multirow{2}{*}{Property}        & \multicolumn{2}{c}{Relative RMSE} & \multirow{2}{*}{\% Improvement} \\
                                 & Our Pipeline   & GP model  &                               \\
                                 \midrule
Hansen Dispersion                & 0.536          & 0.690            & 28.73                              \\

Hansen Polarity                  & 0.509          & 0.605            & 18.86                             \\

Hansen Bonding                   & 0.396          & 0.66            & 66.67                              \\

Dielectric Constant ($\varepsilon_r$) & 0.517          & 0.771            & 48.84                              \\

Gutmann Donor Number                     & 0.458          & 0.865            & 84.83                              \\
Acceptor Number                 & 0.397         & 0.432           & 8.10                             \\
Kamlet-Taft $\beta$              & 0.448          & 0.762            & 68.96                              \\

Kamlet-Taft $\alpha$                     & 0.411               & 0.453          & 9.27                             \\
Kamlet-Taft $\pi*$                  & 0.573         & 0.669            & 14.34                           \\                     
\bottomrule                     
\end{tabular}
\caption{Comparison of Relative RMSE of our model and a gaussian processes model. }\label{tab:compariosn_models}
\end{table}

\FloatBarrier
\subsection*{Application-oriented Results for Solvent Substitution}
To apply the above-developed pipeline for material discovery, we predict solubility parameters on a database of 5000 VOCs\cite{CompsolDatabaseRevised2024}. The final prediction is an average of the predictions of every individual model of the n-fold validation. 
The use of transfer learning from large foundational models increases the chances of good generalization on unseen data. In addition,  the  rigorous cross validation approach makes a strong case for our model's generalization abilities.

\begin{figure}[h]
\centering

\includegraphics[width=1\textwidth]{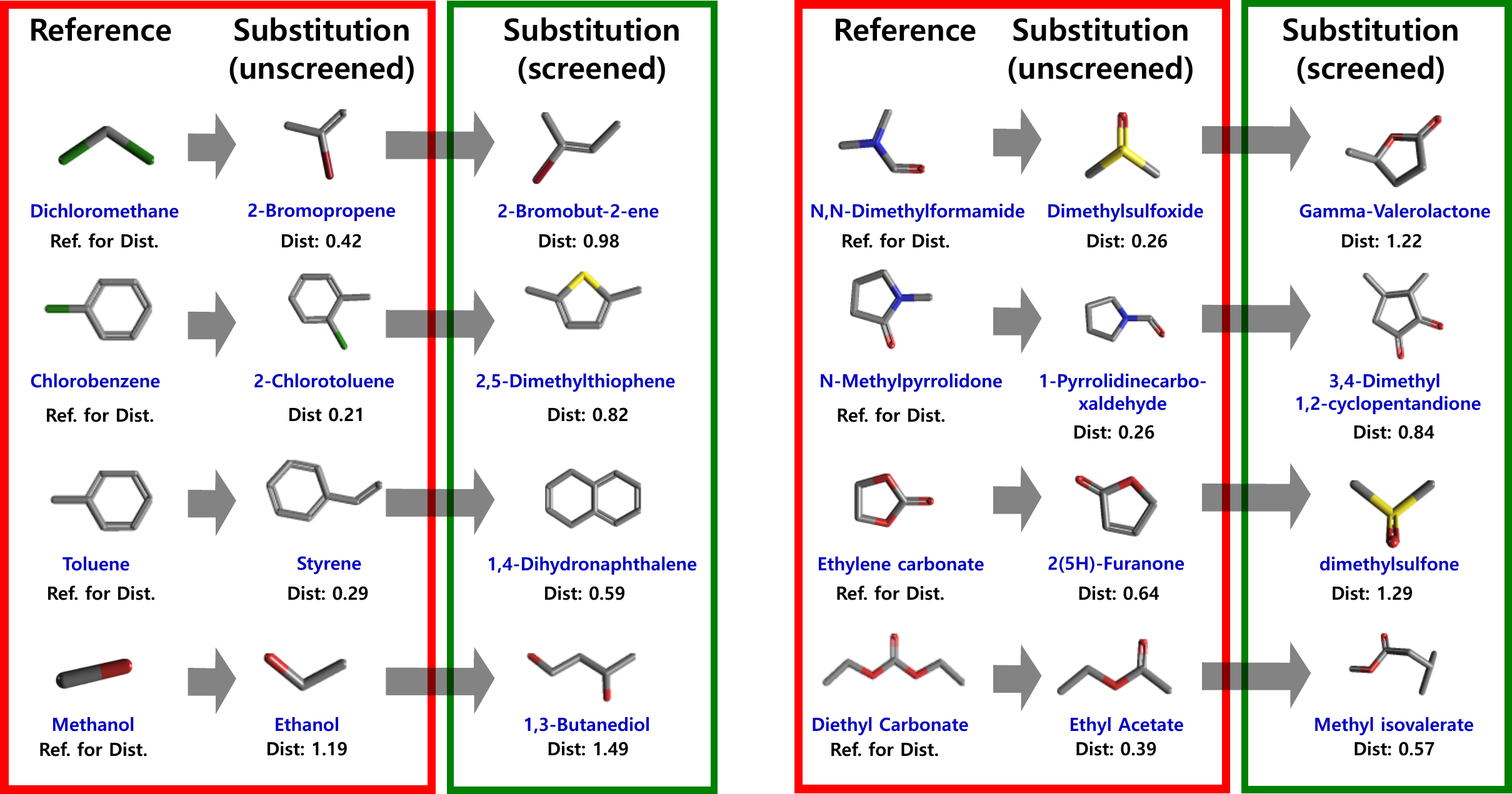}

\caption{Comparison reference solvent molecules with molecules situated at closest Euclidean distance calculated by \cref{eq:dist}. The closest distance increases for the screened results where H3xx and H4xx Ghs-symbols were eliminated. These predictions are merely qualitative plausibility checks. For a real application, the complete list of the n closest predictions should be analyzed by an expert and additional criteria such as reactivity, boiling points, material prices should be considered. }
\label{fig:predictions}
\end{figure}

For qualitative analysis, we applied our substitution tool on a set of VOCs of concern listed in \cref{tab:applications_solvents}---commonly used in a broad range of industrial applications from pharmaceutics, polymer processing and photovoltaics fabrication to battery electrolytes. \cref{fig:predictions}  depicts the molecular structures of these solvents next to the candidates for solvent substitution at closest distance without screening and after screening of H3xx and H4xx symbols, respectively. Note that these closest-distance predictions are prone to random error fluctuations and that we have not considered any other material properties besides solubility parameters such as reactivity, stability, boiling points, rheology, etc.... The predictions are therefore an exemplary "top-1 trial" without any expert knowledge, which is meant as an academic, qualitative plausibility check. In a real-world application, experimentalists should analyze the whole list of n-best candidates for materials discovery, factoring in their chemical expertise and intuition. For this purpose, they can either work directly on the raw predicted data or, more conveniently, use our solvent tool (see \cref{sec:screening}). The tool provides additional degrees of freedom by (i) enabling and disabling screening of certain hazards, (ii) custom weighting of solubility parameters and (iii) propagating prediction uncertainty in the distance metric (Mahalanobis, kl-div) or simply listing it in a separate column (Euclidean). 

Overall, the "top-1 trial" predictions are very plausible: functional groups of reference solvents are frequently represented in the proposed substitute solvents, as well. This is expected due to the correlation between molecular structure and energy interaction properties summarized by solubility parameters. For unscreened predictions, many are well aligned with chemical intuition, advocating for the correctness of the employed methodology. For instance, ethanol is predicted to have the most similar solubility behavior as methanol, 2-chlorotoluene as chlorobenzene, dimethylsulfoxide as N,N-dimethylformamide and ethyl acetate as diethyl carbonate. However, since we did not consider the reactivity, other predictions (styrene, 1-pyrrolidinecarboxaldehyde and 2(5H)-furanone) are likely not of industrial relevance. Still, we included them here to underscore that we did not bias, filter or reduce our data to prioritize plausible results.  

\begin{table}[ht]
\centering
\label{tab:solvent_substitutions}
\renewcommand{\arraystretch}{1.4}
\begin{tabular}{p{4.5cm}p{9.5cm} p{1.5cm}}
\hline
\textbf{Solvent} (Substitue) & \textbf{Key Industrial Applications} & \textbf{Ref.} \\
\hline
\textbf{Dichloromethane } \linebreak (2-Bromobut-2-ene) & Pharmaceuticals, paints, polymer films, agrochemicals, extraction  & \cite{byrneToolsTechniquesSolvent2016} \\
\hline
\textbf{Chlorobenzene}  \linebreak(2,5-Dimethylthiophene) & Hybrid and organic solar cells, polymers, dyes, agrochemicals, rubber chemicals & \cite{capelloWhatGreenSolvent2007} \\
\hline
\textbf{Toluene}  \linebreak (1,4-
Dihydronaphthalene) & Paints \& coatings, adhesives, rubber processing, pharmaceuticals, natural product extraction  & \cite{sheldonGreenSolventsSustainable2005, pellisSaferBiobasedSolvents2019} \\
\hline
\textbf{Methanol}  \linebreak (1,3-Butanediol) & Biodiesel production; pharmaceuticals; natural product extraction; fuel cells & \cite{erchamoImprovedBiodieselProduction2021, leeInfluenceSolventChoice2024} \\
\hline
\textbf{\textit{N,N}-Dimethylformamide}  \linebreak ($\gamma$-Valerolactone) & Hybrid and organic solar cells, polyamide/polyurethane processing, pharmaceuticals, electrochemistry, synthetic fibers &  \cite{heraviSolventTripleRoles2018} \\
\hline
\textbf{\textit{N}-Methyl-2-pyrrolidone}   \linebreak (3,4-Dimethyl 1,2-cyclopentandione) & Li-ion battery electrodes; membranes; pharmaceuticals; paints  & \cite{basmaLocalStructurePolar2018, NewSolventSeeks} \\
\hline
\textbf{Ethylene$\,\,$carbonate} \linebreak (Dimethylsulfone)  & Li-ion battery electrolytes, polycarbonate synthesis, epoxies & \cite{chenTraceEthyleneCarbonatemediated2024, westphalTemperatureResolvedCrystalStructure2025} \\
\hline
\textbf{Diethyl$\,\,$carbonate}  \linebreak  (Methyl isovalerate) & Li-ion battery electrolytes; polycarbonate synthesis; carbonylation; fuel oxygenate; plasticiser & \cite{joteEffectDiethylCarbonate2020, shuklaDiethylCarbonateCritical2016} \\
\hline
\end{tabular}
\caption{Common industrial solvents and proposed substitutes with typical applications.}\label{tab:applications_solvents}
\end{table}

We now continue the qualitative analysis of our "top-1 trial" predictions for the results screened by H3xx and H4xx symbols. Dichloromethane is a widely used solvent in the chemical industry, which is challenging to subsitute due to its high solvation ability and high volatility. Our tool proposes 2-bromobut-2-ene as a candidate for substitution. While this is still a halogenated chemical that is not considered ``green'' in a strict definition it passes the rudimentary screening applied herein. It is a sensible suggestion given the highly halogenated dichloromethane.  However, the functionality of 2-bromobut-2-ene as a solvent is questionable due to its reactivity\cite{orkinPhotochemistryBromineContainingFluorinated2002}. (The next closest candidate is food agent  2,5-dimethylthiophene just as suggested for substitution of Chlorobenzene.) The proposed "green" substitution of the aromatic, high-boiling point solvents,  Chlorobenzene and Toluene are 2,5-dimethylthiophene and 1,4-dihydronaphthalene, preserving (partially) the aromatic ring motif. We have not found any instance of them being used as a solvent before, which is most likely not practical due to their reactivity. The widely employed, polar protic solvent methanol is proposed to be replaced by 1,3-butanediol, which has been employed as a solvent before\cite{romeroFeasibility13butanediolSolvent2010} (but has a significantly higher boiling point). The high-bioling point, polar aprotic solvent  N,N-dimethylformamide is proposed to be substituted by gamma-Valerolactone (GVL), which is perfectly consistent with green solvent fabrication of perovksite solar cells\cite{hanGreenSolutionProcessing2024,kimEcoFriendlyAllLayerGreen2024,baeAdvancingPerovskiteSolar2025}. N-methyl-2-pyrrolidone is proposed to be replaced by the cyclic diketone 1-methyl-2,3-cyclohexadione used in the flavor industry.  Ethylene carbonate, which is heavily used as a lithium-ion battery electrolyte, is proposed to be substituted by dimethylsulfone which has been demonstrated successfully as a battery electrolyte before\cite{hofmannNovelElectrolyteMixtures2015}. The prediction is therefore consistent with literature. Diethyl carbonate is suggested for substitution with the flavor enhancing ester methyl iosvalerate, which adds to the more exotic predictions of the previously reported 1-methyl-2,3-cyclohexadione. While we could not find any instance of these exotic molecules for use as a solvent and economical feasibility is questionable, exotic predictions are expected when employing our pipeline for "top-1 trial" predictions for material discovery. As stated above, they underline that we did not filter, bias or reduce our data.

We sum up that our tool is very useful for data-informed decisions of solvent substitution for experimentalists because several predictions match well with solvents that have been used as green alternatives before. We would like to repeat that we have only shown the closest distance suggestions here, expecting a certain rate of exotic predictions due to the methodology. Most strikingly, the tool succeeds in predicting three green solvent/electrolyte alternatives that were already used in literature, underlining the consistency of the applied methodology.


\FloatBarrier

\section{Conclusion and Outlook}

In this work, we present a versatile machine learning pipeline based on transfer learning from the foundational model DimeNet with optional Attention used in conjunction with molecular descriptors. The final layer is a Gaussian Processes model to achieve a data efficient, uncertainty aware, highly robust and accurate framework. We applied this pipeline to the prediction of solubility parameters and the determination of novel, efficient and green solvents. We demonstrate that our pipeline has a good performance not only for the large datasets of the three Hansen solubility parameters and Dielectric constants. It also reaches a good performance on datasets of several hundreds and even less than 50 datapoints for predicting Gutmann Acceptor/Donor numbers as well as Kamlet-Taft solvatochromic parameters. These characteristics uniquely underline the benefits  of transfer learning leveraging pre-trained foundational models. For creating a customizable solvent screening tool, our model is applied to the CompSol database of over 5000 volatile organic compounds, augmenting the available datasets for the given parameters by a factor of 4 for the Hansen parameters, and up to a factor of 100 for other solubility parameters. Since extrapolation is a hard claim, especially for small datasets, we have performed strict 5- to 10-fold validation to avoid presenting unrepresentative optimistic results. In the validation process, our model's performance remained consistently accurate, giving us confidence in the extrapolative power of our results.  Notably, the tool is deployed as an easy-to-use web app for chemists and engineers who are not machine learning experts (see \url{https://kouter.streamlit.app/}). The tool includes customization of distance weighting, uncertainty filtering and  
screening filters for eliminating solvents of concern\footnote{the tool is for discovery of new chemicals only and should not be employed for any safety-related decisions}.  On top of that, the codes for the ML process are fully disclosed as well as the obtained, predicted data-sets. In this way, we provide the unique opportunity to the community to re-use our methods and data for their particular application, emphasizing the relevance and impact of this work. An additional benefit of leveraging pre-trained foundational models is that the pipeline is very lightweight - deployment on personal computers is feasible. This further underlines the importance of this work, making it available even to groups without specialized computational infrastructure. In this context, we would like to stress as well that DimeNet is already integrated in the PyTorch library. Since its release however, it has been outperformed by other, more intricate models, notably, MACE and NequIP. 

In the future, the DimeNet model will be substituted by more recently developed foundational models such as NequIP or MACE as the pipeline is fully adaptable. In addition, even more solubility parameters such as the Catalàn parameters or Fawcett Acidity and Basicity will be predicted. Since the pipeline is very adaptable in its target range, it could also be used to predict other functions of chemicals where available data is sparse such as charge transport properties for electronics or reaction pathways. Finally, instead of using a database of known liquid molecules, new molecules that have never been synthesized could be investigated.  The present work should therefore be seen as a first, powerful proof-of-concept demonstration of what this pipeline is able to achieve. As a part of this demonstration, the tool released with the publication will help emerging energy technology roll-out to become more eco-friendly by substituting solvents of concern. In the future, this pipeline can be coupled with high throughput labs to efficiently expedite material discovery and development.


\section*{Data Availability}
upon publication
\section*{Code Availability}
upon publication




\section*{Credits}
The author contributions according to the CRediT taxonomy can be seen in \cref{tab:credit}.

\begin{table}[ht]
\centering
\caption{Author Contributions according to the CRediT taxonomy}
\label{tab:credit}
\begin{tabular}{p{4.5cm}p{10cm}}
\hline
\textbf{Contribution} & \textbf{Author(s)} \\
\hline
Conceptualization & Aldo Di Carlo, Simon Ternes, Ioannis Kouroudis,  Alessio Gagliardi \\
Methodology & Ioannis Kouroudis, Simon Ternes \\
Software & Ioannis Kouroudis, Simon Ternes, Zhaosu Gu , Gohar Siddiqui \\
Validation & Ioannis Kouroudis \\
Formal analysis &  Ioannis Kouroudis,Simon Ternes, Gohar Siddiqui\\
Investigation & Simon Ternes, Ioannis Kouroudis, Marina Ustinova, Angelo Lembo, Alessio Gagliardi, Aldo Di Carlo \\
Resources & Ioannis Kouroudis, Simon Ternes, Alessio Gagliardi \\
Data curation & Simon Ternes \\
Writing -- Original Draft & Simon Ternes, Ioannis Kouroudis \\
Writing -- Review \& Editing & Ioannis Kouroudis, Simon Ternes, Gu Zhaosu, Gohar Siddiqui, Marina Ustinova, Angelo Lembo, Aldo Di Carlo, and Alessio Gagliardi \\
Visualization & Simon Ternes, Ioannis Kouroudis \\
Supervision & Simon Ternes,  Ioannis Kouroudis\\
Project administration & Aldo Di Carlo, Alessio Gagliardi \\
Funding acquisition &  Aldo Di Carlo, Alessio Gagliardi  \\
\hline
\end{tabular}
\end{table}

\FloatBarrier
\section*{ACKNOWLEDGEMENTS}
 
  A.G. acknowledges financial support from TUM Innovation Network for Artificial Intelligence powered Multifunctional Material Design (ARTEMIS) and funding in the framework of Deutsche Forschungsgemeinschaft (DFG, German Research Foundation) under Germany's Excellence Strategy – EXC 2089/1 – 390776260 (e-conversion).

 S.T. acknowledges the European Union's Framework Programme for Research and Innovation Horizon Europe (2021-2027) under the Marie Sklodowska-Curie Grant Agreement No. 101107885 “INT-PVK-PRINT”. S.T. and A.L. acknowledge the European Union's Framework Programme for Research and Innovation Horizon Europe (2021-2027) under the Grant Agreement No. 101147311 “LAPERITIVO”.

A.D.C. acknowledges the European Union’s Framework Programme for Research and Innovation Horizon
Europe (2021-2027) under the Grant Agreement No. 101122327 “SMARTLINE-PV”.

M. U. gratefully acknowledges financial support of the GoPV project of "Bando di gara per progetti di ricerca di cui all'art. 10, comma 2, lettera a) 26/10/2000 Piano triennale 2019–2021 della Ricerca di sistema elettrico nazionale" (n. CSEAA\_00011)

 This project has received funding from the MENTOR, European Union’s Horizon Europe (HORIZON) Marie Sklodowska-Curie Actions Doctoral Networks (MSCA-DN) HORIZON-MSCA-2023-DN-01 call, under the Grant Agreement number 101169056. The project duration is from 1 October 2024 to 30 September 2028.

\FloatBarrier

\printbibliography
\end{document}